\documentclass[letterpaper, 10 pt, conference]{ieeeconf}  

\IEEEoverridecommandlockouts                              

\usepackage{graphics} 
\usepackage{epsfig} 
\usepackage{mathptmx} 
\usepackage{times} 
\usepackage{amsmath} 
\usepackage{amssymb}  
\usepackage[dvipsnames]{xcolor}
\usepackage{booktabs}
\usepackage{multirow}
\usepackage{makecell}
\usepackage{pifont} 
\usepackage{colortbl}
\usepackage{amssymb}
\newcommand{\cmark}{\checkmark}
\definecolor{darkgreen}{RGB}{59,125,35}
\definecolor{light_purple}{RGB}{216,110,204}
\definecolor{dark_blue}{RGB}{8, 79, 106}
\definecolor{dark_purple}{RGB}{102,0,102}
\definecolor{Grey}{rgb}{0.5, 0.5, 0.5}

\title{\LARGE \bf

SmartWay: Enhanced Waypoint Prediction and Backtracking for Zero-Shot Vision-and-Language Navigation
}

\author{Xiangyu Shi$^*$, Zerui Li$^*$, Wenqi Lyu, Jiatong Xia, Feras Dayoub, Yanyuan Qiao$^\mathsection$, Qi Wu$^\dagger$
\thanks{The authors are with the Australian Institute for Machine Learning at the University of Adelaide, Adelaide, Australia.}%
\thanks{* These authors contributed equally}
\thanks{$\mathsection$ Project lead: Yanyuan Qiao (\tt\footnotesize yanyuan.qiao@adelaide.edu.au)}
\thanks{$\dagger$ Corresponding author: Qi Wu ({\tt\footnotesize qi.wu01@adelaide.edu.au})}
}

\begin{document}

\maketitle
\thispagestyle{empty}
\pagestyle{empty}

\begin{abstract}
Vision-and-Language Navigation (VLN) in continuous environments requires agents to interpret natural language instructions while navigating unconstrained 3D spaces. Existing VLN-CE frameworks rely on a two-stage approach: a waypoint predictor to generate waypoints and a navigator to execute movements. However, current waypoint predictors struggle with spatial awareness, while navigators lack historical reasoning and backtracking capabilities, limiting adaptability. We propose a zero-shot VLN-CE framework integrating an enhanced waypoint predictor with a Multi-modal Large Language Model (MLLM)-based navigator. Our predictor employs a stronger vision encoder, masked cross-attention fusion, and an occupancy-aware loss for better waypoint quality. The navigator incorporates history-aware reasoning and adaptive path planning with backtracking, improving robustness. Experiments on R2R-CE and MP3D benchmarks show our method achieves state-of-the-art (SOTA) performance in zero-shot settings, demonstrating competitive results compared to fully supervised methods. Real-world validation on Turtlebot 4 further highlights its adaptability.
\end{abstract}

\section{Introduction}
\label{sec:intro}

Vision-and-Language Navigation (VLN) is a complex, cross-disciplinary challenge that entails interpreting natural language instructions and navigating unseen environments through sequential decision making~\cite{anderson2018vision, qi2020reverie, ku2020room, hong2022bridging, wang2023scaling}. While most VLN research has been conducted in discrete environments, recent efforts have shifted toward more realistic continuous settings, which are often termed Vision-and-Language Navigation in Continuous Environments (VLN-CE), to better capture the complexities of real-world navigation. Unlike discrete settings, VLN-CE removes the constraints of predefined navigation graphs, requiring the agent to execute low-level actions and contend with dynamic uncertainties~\cite{an2022bevbert, wang2023gridmm, an2023etpnav}. This transition demands more flexible navigation strategies, as agents must infer traversable paths without relying on fixed nodes or edges. To handle these challenges, mainstream VLN-CE approaches employ a two-stage framework: (1) Waypoint Predictor to identify navigable locations, followed by~\cite{hong2022bridging} (2) Navigator to make action decisions and execute movements based on these waypoints.

A waypoint predictor generates waypoints directly from RGB-D observations, enabling flexible and adaptive navigation. Unlike discrete VLN methods that rely on predefined navigation graphs, this approach allows agents to make fine-grained movement decisions without being constrained by fixed nodes or edges.
At each step, the agent captures panoramic RGB and depth images, encodes them using separate vision backbones, and fuses the extracted features into a unified representation. A lightweight Transformer then models spatial relationships to predict waypoints, facilitating efficient exploration of new environments. By dynamically selecting waypoints, the waypoint predictor enhances navigation efficiency and adaptability to unseen environments, making it a crucial component for real-world deployment. Despite its importance, existing waypoint predictor frameworks have several limitations. The most widely used model, proposed by Hong et al.~\cite{hong2022bridging}, employs a ResNet-50~\cite{resnet} encoder pretrained on ImageNet~\cite{deng2009imagenet} to process RGB images, along with a simple linear layer for RGB-D fusion. However, recent studies~\cite{an2023etpnav} indicate that RGB information contributes little to feature expressiveness in waypoint prediction, limiting the model's generalizability. In addition, the waypoint predictor lacks explicit environmental constraints, making it susceptible to predicting waypoints that lead to obstacles or suboptimal paths.

\begin{figure}[t]
    \centering
    \includegraphics[width=.99\linewidth]{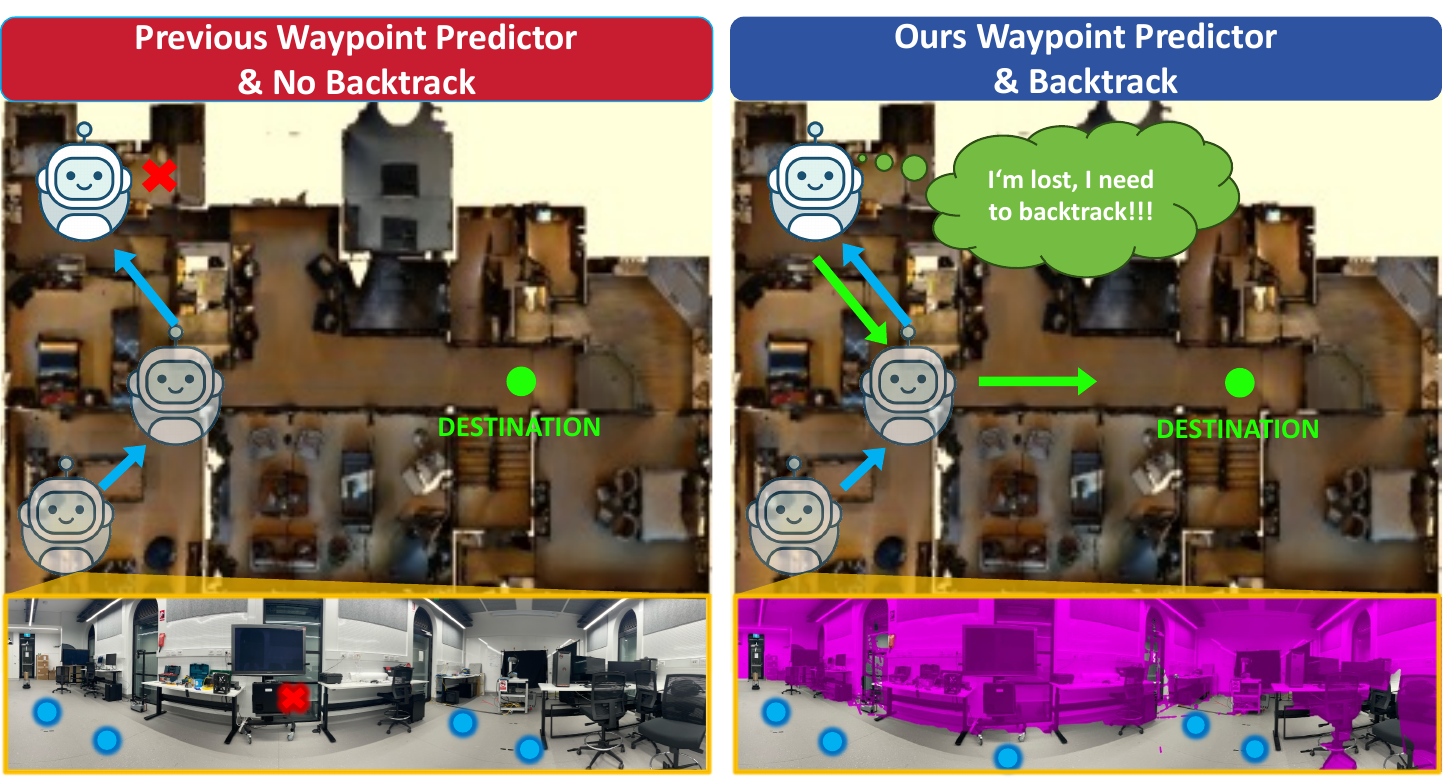}

   \vspace{-10pt}
    \caption{Role of our proposed waypoint predictor and backtrack mechanism. In the \textbf{No Backtrack} scenario (left), the agent gets lost 
    due to irreversible errors. In contrast, the \textbf{Backtrack-enabled} agent (right) detects a navigational failure, retraces its steps, and finds an alternative path, enhancing robustness in complex environments. In addition, our enhanced waypoint predictor is notably more effective and reliable in open spaces due to the effect of occupancy loss.
}
    \label{fig:teaser}
    \vspace{-20pt}
\end{figure}

To address these limitations, we introduce three key improvements to waypoint predictor: (1) a stronger vision encoder, replacing ResNet-50 with DINOv2~\cite{dinov2} to capture richer scene representations; (2) a masked cross-attention fusion mechanism, which enhances RGB-D feature interaction and spatial awareness; and (3) an occupancy-aware loss, ensuring that predicted waypoints align with navigable regions. These enhancements improve the robustness and spatial reasoning of waypoint predictor, leading to more effective navigation in complex, real-world environments.

Despite advancements in waypoint prediction, real-world deployment remains challenging due to data scarcity and limited generalization~\cite{wang2023scaling}. Large language models (LLMs) offer a potential solution by leveraging large-scale web data, potentially reducing reliance on domain-specific annotations and improving generalization. However, existing LLM-based navigation methods struggle with continuous motion control and obstacle avoidance, as they are primarily designed for discrete environment settings like NavGPT, MapGPT~\cite{zhou2024navgpt,long2023discuss,dai2023instructblip}. Recent approaches have explored LLM-driven navigation in VLN-CE, particularly in zero-shot settings where the agent generalizes without environment-specific fine-tuning~\cite{ca-nav,opennav}. Open-Nav~\cite{opennav} represents a recent attempt to address these challenges by leveraging open-source LLMs for zero-shot VLN-CE. However, these LLMs rely exclusively on textual inputs and lack the capacity to directly interpret visual information. In such systems, visual perception is abstracted by converting RGB images into textual descriptions by Vision-Language Models 
as an intermediary representation, resulting in an indirect and incomplete understanding of the visual environment. In addition, prior work in learning-based VLN has highlighted the critical importance of effectively utilizing navigational history and incorporating backtracking mechanisms~\cite{an2022bevbert, An2023ETPNavET, chen2022think}. The reliance on textual conversion for historical observations prevents the full exploitation of rich contextual information and cannot effectively backtrack. 

Therefore, to overcome these limitations, we propose an MLLM-based VLN agent that (1) integrates historical trajectory reasoning through a \textit{History-aware Single-expert Prompt System}, and (2) improves adaptability with an \textit{Adaptive Path Planning} module with backtracking capabilities, ensuring robust navigation in zero-shot settings.

In our experiments, we evaluate our approach in both simulated and real-world environments. In the simulation, we test on R2R-CE val-unseen set within the Habitat simulator. Our zero-shot navigator achieves a success rate (SR) of 29\% and an SPL of 22.46\%, outperforming all other zero-shot methods. In real-world experiments, we deploy our method on a Turtlebot 4 equipped with an OAK-D Pro camera, conducting navigation tasks across 25 diverse instructions. Our method surpasses learning-based baselines, demonstrating improved navigation efficiency and robustness. Additionally, our proposed backtrack mechanism further enhances success rates and reduces navigation errors, confirming its effectiveness in mitigating compounding mistakes.

In this work, our main contributions are as follows:
\begin{itemize}
    \item We propose an enhanced waypoint predictor module that improves accuracy by leveraging a robust vision encoder, a masked cross-attention fusion strategy, and an occupancy-aware loss function, all tailored for continuous navigation settings.
    \item We conduct the first exploration of Multimodal Large Language Models (MLLMs) for VLN in continuous environments under a zero-shot paradigm, introducing a History-aware Single-expert Prompt System to effectively integrate past trajectory information and enhance navigational reasoning.
    \item We introduce a novel backtracking mechanism for MLLM-based VLN agents in continuous spaces, and empirically demonstrate its critical role in mitigating error propagation and enhancing overall navigation performance.
\end{itemize}

\section{Related Work}
\label{sec:related work}
\subsection{Vision-and-Language Navigation(VLN)}

Vision-and-Language Navigation (VLN) is a task where an agent learns to follow human instructions to navigate in previously unseen environments~\cite{anderson2018vision,zhang2024vision}. Most existing VLN benchmarks are built on discretized simulated scenes with predefined navigation graphs~\cite{qi2020reverie, anderson2020rxr, thomason2020cvdn}. To bridge the gap between simulation and real-world applications, Krantz et al.~\cite{krantz2020beyond} introduced a benchmark that extends the VLN task from discrete to continuous environments (CE), making it more representative of real-world scenarios. In efforts to boost performance in VLN-CE, researchers have adopted cross-modal alignment strategies from discrete settings and proposed waypoint models aimed at bridging the gap between traditional VLN and continuous navigation scenarios such as CMA, RecBERT and ETPNav~\cite{hong2022bridging,krantz2020beyond,an2023etpnav}. Nonetheless, the decision-making process of VLN policies remains heavily contingent on generating waypoints, which are frequently misaligned with the intended sub-goals. Moreover, beyond the challenges of accurate waypoint prediction, the deployment of these systems on physical robotic platforms continues to exhibit limited generalization. 
Therefore, we aim to enhance waypoint prediction and leverage MLLMs’ strong multimodal reasoning capabilities, integrating visual and action-oriented modalities. Our goal is to explore training-free VLN in continuous environments, employing MLLMs as zero-shot navigators.

\subsection{Waypoint Prediction}

In VLN-CE, agents navigate a continuous 3D space rather than relying on a fixed navigation graph as in discrete VLN. Without the benefit of discrete nodes, agents must dynamically predict navigable waypoints. Hong et al.~\cite{hong2022bridging} introduced a waypoint predictor that generates candidate waypoints during navigation, however, this approach’s heavy reliance on spatial cues often yields waypoints that do not align well with the intended sub-goals. To address this limitation, Dong et al.~\cite{an2022bevbert, an2023etpnav} improve waypoint placement accuracy by training exclusively on depth images in open spaces, arguing that RGB information provides limited enhancement in feature expressiveness and generalization in indoor environments. Furthermore, Li et al.~\cite{li2025ground} augment the training datasets to reinforce spatial priors and capture a more comprehensive representation of real-world scenarios, but still maintained the original network architecture. Unlike prior approaches that fail to effectively utilize RGB features or lack explicit spatial constraints, which results in suboptimal waypoint placement, our work enhances waypoint prediction by incorporating a stronger vision encoder (DINOv2~\cite{dinov2}), a masked cross-attention fusion mechanism for RGB-D interaction, and an occupancy-aware loss. These improvements explicitly model spatial constraints and enhance generalization across diverse navigation scenarios.

\subsection{Foundation Models as Embodied Agents}
In recent years, researchers have increasingly integrated foundation models into various embodied domains, such as VLN tasks, revealing substantial potential in utilizing these models as effective navigation agents \cite{zhou2024navgpt, long2023discuss, liang2023mo, schumann2024velma, zhang2024navid, qiao2023march}. In light of this, a growing body of research has leveraged the strong generalization capabilities of LLMs to enhance VLN performance, either by examining the inherent navigational reasoning capacities of LLMs, or by integrating LLMs into navigation systems through modular frameworks~\cite{zhou2024navgpt, long2023discuss,  opennav, pan2023langnav, zheng2024towards, chen2024mapgpt}. For example, MapGPT~\cite{chen2024mapgpt} constructs a topological map to enable GPT to comprehend global environments, limited to discrete settings, while DiscussNav~\cite{long2023discuss} employs a multi-expert framework to delegate different navigation tasks to GPT. However, these methods largely rely on converting visual and spatial information into textual prompts, which limits their effectiveness in representing complex spatial relationships~\cite{opennav, tan2019lxmert}. Moreover, although history and backtracking mechanisms have proven effective in learning-based VLN-CE approaches~\cite{an2023etpnav, chen2022think,chen2021history}, their critical importance when employing MLLMs as navigators has not been thoroughly investigated. In response, this work aims to advance VLN by leveraging MLLM as navigating agent. Specifically, we explore the performance gains afforded by the incorporation of rich historical context and backtracking functionality in VLN-CE, thereby establishing a more robust framework for VLN.

\section{PRELIMINARIES}

\subsection{Problem Formulation}
\label{sec:problem_formulation}

In this work, we address \textit{Vision-and-Language Navigation in Continuous Environments} (VLN-CE), wherein an autonomous agent must navigate within a continuous 3D space \(\mathbf{E}\) based on linguistic instructions. Let \(\mathbf{x}_{t} = (x_{t},y_{t},z_{t}) \in \mathbf{E}\) denote the agent’s position at time \(t\). The agent receives panoramic RGBD observations collected at evenly spaced viewpoints (e.g., \(0^\circ, 30^\circ, \dots, 330^\circ\)), yielding 12 RGB and 12 depth images \(I = \{ (I^{rgb}_i, I^{depth}_i) \mid i = 1, \dots, 12 \}\), where \( I^{rgb}_i \in \mathbb{R}^{H \times W \times 3} \) and \( I^{depth}_i \in \mathbb{R}^{H \times W} \). Currently, the agent receives instruction \(L = \{l_1, l_2, \dots, l_n\}\), where each \(l_i\) is a token (word) of the instruction. The instruction specifies how to reach a goal location \(\mathbf{x}_{\text{goal}} \in \mathbf{E}\) starting from \(\mathbf{x}_{\text{start}} \in \mathbf{E}\). The navigation process unfolds via discrete low-level actions that dictate movement direction and distance. By executing these actions in accordance with the instructions, the agent aims to minimize navigation errors and arrive at the specified target in the continuous 3D environment.

\subsection{Waypoint Predictor}

The waypoint predictor bridges the gap between discrete environments and continuous space by generating candidate waypoints based on the current observation. For any given position in an open environment, the agent captures RGB and depth panoramas, each consisting of 12 single-view images captured at 30-degree intervals. These images are encoded into feature sequences, denoted as $F^{rgb} = \{ f^{rgb}_i \mid i = 1,\dots,12 \}$ and $F^{depth} = \{ f^{depth}_i \mid i = 1,\dots,12 \}$. RGB images are encoded using a ResNet-50~\cite{resnet} pre-trained on ImageNet~\cite{deng2009imagenet}, while depth images are encoded using another ResNet-50 pre-trained for point-goal navigation~\cite{wijmans2020ddppo}. Each feature pair $(f^{rgb}_i, f^{depth}_i)$ is fused using a non-linear layer $W^{m}$ to obtain a combined feature, $f^{rgbd}_i$. A two-layer Transformer processes the 12 fused features to analyze spatial relationships and predict adjacent waypoints.

Each single-view image is square and has a 90-degree field of view (FoV), covering three 30-degree sectors. To maintain spatial consistency, self-attention for each $f^{rgbd}_i$ is restricted to its adjacent features. The Transformer outputs $\tilde{f}^{rgbd}_i$, which encodes spatial relationships centred on image $i$. A classifier generates a $120 \times 12$ heatmap $\mathbf{P}$, where angles are spaced every 3 degrees, and distances range from 0.25m to 3.00m in 0.25m steps. Non-maximum suppression (NMS) is then applied to extract the top $K$ navigable waypoints, and $K$ is set up to 5.
The waypoint predictor is trained using a regression loss that minimizes the difference between the predicted heatmap $\mathbf{P}$ and the ground-truth heatmap $\mathbf{P}^*$. Specifically, we use a mean squared error (MSE) loss formulated as:  
\begin{equation}
L_{vis} = \text{MSE}(\mathbf{P}, \mathbf{P}^*)
\label{eq:L_vis}
\end{equation}

where $\mathbf{P}$ is the predicted waypoint probability distribution, and $\mathbf{P}^*$ is the ground-truth heatmap. This loss ensures that the model accurately predicts the spatial distribution of navigable waypoints.

\section{Method}
\label{sec:method}

\begin{figure*}
    \centering
    \includegraphics[width=0.85\linewidth]{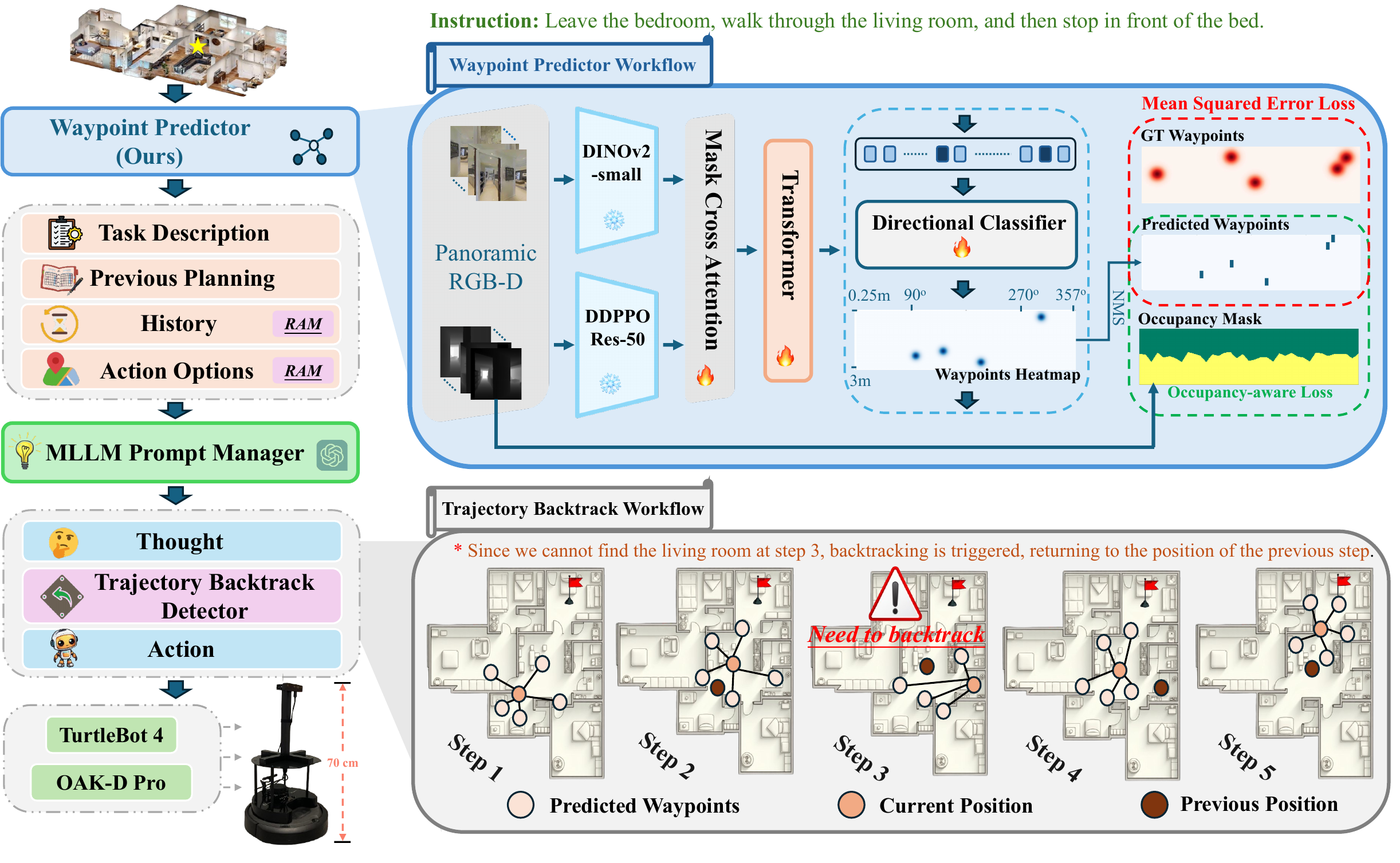}
    \vspace{-15pt}
    \caption{Our approach consists of two key components: an Occupancy-aware Waypoint Predictor and an MLLM-based Navigator. The waypoint predictor refines waypoint selection by integrating a stronger vision encoder, a masked cross-attention fusion mechanism, and an occupancy-aware loss, improving prediction quality. The MLLM-based Navigator processes candidate waypoints using visual and textual information to enhance navigation decisions, incorporating finer turning options, historical context, and a backtracking strategy. The robot on this figure is our Turtlebot 4 mobile robot equipped with an OAK-D Pro camera mounted at a height of 70 cm.}
    \label{fig:overview}
    \vspace{-15pt}
\end{figure*}

\begin{figure}
    \centering
    \includegraphics[width=\linewidth]{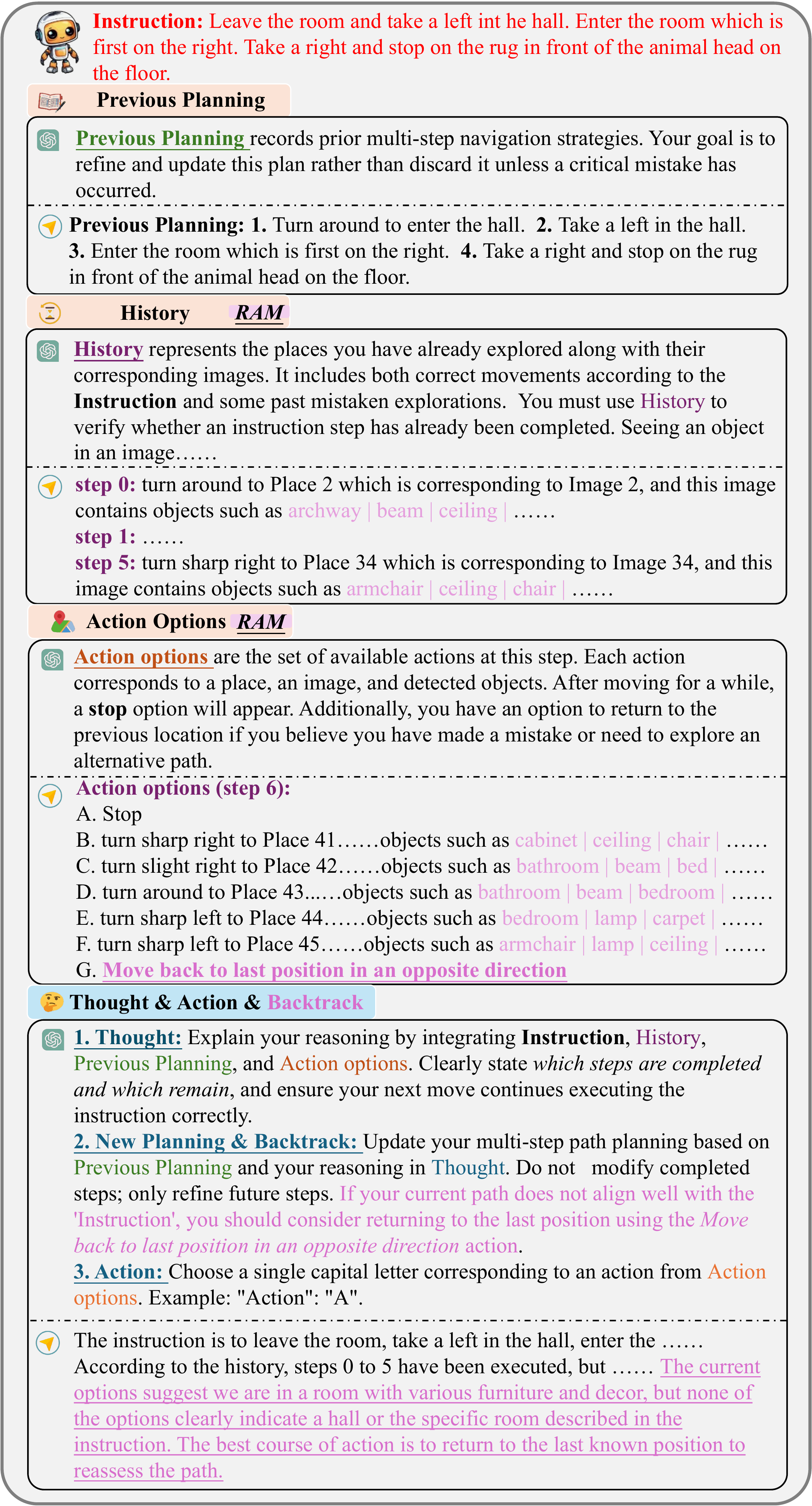}
    \vspace{-20pt}
    \caption{Given an \textcolor{red}{instruction}, the MLLM first checks whether there is a \textcolor{darkgreen}{Previous Planning}, then reviews the explored \textcolor{dark_purple}{History}. Based on the \textcolor{orange}{Current Action Options,} it generates a \textcolor{dark_blue}{reasoning process, updates the plan, and selects the next action}. Notably, a \underline{\textcolor{light_purple}{backtrack policy}} allows the robot to return to previously visited locations when necessary.}
    \label{fig:prompt}
    \vspace{-15pt}
\end{figure}

As shown in Fig~\ref{fig:overview}, our approach consists of two key components: an Occupancy-aware Waypoint Predictor and an MLLM-based Navigator. The waypoint predictor (Sec.~\ref{sec:waypoint_predictor}) refines waypoint selection by integrating a stronger vision encoder, masked cross-attention fusion, and an occupancy-aware loss, improving prediction accuracy and navigability in continuous spaces. Then, the MLLM-based Navigator (Sec.~\ref{sec:method_nav}) processes candidate waypoints using both visual and textual information to improve decision-making. It enhances navigation by adding finer turning options, keeping track of past actions, and using a backtracking strategy to correct mistakes and avoid unnecessary exploration.

\subsection{Occupancy-aware Waypoint Predictor}
\label{sec:waypoint_predictor}

\subsubsection{Vision Encoder Upgrade}

The original waypoint predictor uses a ResNet-50-based RGB encoder \({E}^{resnet}\), pre-trained on ImageNet, to extract visual features from each RGB image \( I^{rgb}_i \) (\( i = 1, \dots, 12 \)), producing embeddings \( f^{rgb}_i = {E}^{resnet}(I^{rgb}_i) \). 
To improve visual representation quality, we replace ResNet-50 with DINOv2~\cite{dinov2}, a self-supervised vision transformer that offers richer and more transferable features. The new RGB encoder \({E}^{dino}\) generates embeddings as \( f^{rgb}_i = {E}^{dino}(I^{rgb}_i) \), while the depth encoder \({E}^{depth}\) remains unchanged, encoding depth images \( I^{depth}_i \) as \( f^{depth}_i = {E}^{depth}(I^{depth}_i) \).  
We chose DINOv2 Small which has a similar size to the ResNet-50 to balance efficiency and expressiveness, achieving improved feature quality without increasing computational cost.
\subsubsection{Masked Cross-Attention for Feature Fusion}
In the baseline approach, RGB and Depth feature $(f^{rgb}_i, f^{depth}_i)$ are simply fused using a non-linear layer $W^m$.
However, such naive fusion does not fully exploit the complementary nature of RGB and depth modalities~\cite{an2023etpnav}. Specifically, RGB provides detailed textures and semantics but lacks depth information, while depth offers spatial structure but lacks semantic details. A simple linear fusion fails to combine them effectively, leading to weak representations.

To improve feature interaction and enhance spatial awareness, we replace $W^m$ with a {masked cross-attention}, similar as the original waypoint predictor's transformer block to improve feature interaction and enhance spatial awareness.

In the cross-attention mechanism, $f^{rgb}_i$ serve as \textit{queries}, while $f^{depth}_i$ act as \textit{keys} and \textit{values}, enabling RGB to focus on crucial depth regions. The updated representation is computed as:
\begin{equation}
\boldsymbol{f}^{rgbd}_{i} = \text{CrossAttention}(\boldsymbol{f}^{rgb}_{i}, \boldsymbol{f}^{d}_{i}, \text{mask})
\end{equation}
The mask here is to maintain spatial consistency, each $f^{rgbd}_i$ is restricted to its adjacent features. 

\subsubsection{Occupancy-aware Loss}
Existing waypoint predictors rely only on ground truth heatmaps for training, ignoring environmental constraints. This can result in waypoints being placed in obstructed areas, making them inaccessible. To address this, we introduce an occupancy-aware loss incorporating depth-based constraints, ensuring waypoints are placed in navigable regions and improving navigation ability.

Having the captured depths $ \{ I^{depth}_i \mid i = 1, \dots, 12 \}$ of each position, we can calculate the shortest relative distance of each interval and the agent's position on the horizontal plane.
Specifically, because the HFOV of each $  I^{depth}_i$ is 90, we uniformly select four captured depths covering a 360-degree field of view, and extract the minimum depth values at evenly spaced columns from the original depth map of size $w \times h$, forming a reduced each depth map of size $30 \times h$ to represent the depth value of each 90 degrees.
And for each column $n$ of the reduced depth map, we extract the pixel $(n,m)$ where $n\in [1,30], m \in [1,h] $ with the minimum depth value, and calculate its 3D coordinates ${C}_{n}=(x_{n}, y_{n}, z_{n})$ with respect to the agent’s position as the origin, which can be formulated as:
\begin{equation}
{C}_{n} = I^{depth}(n,m)\cdot{d}_{n,m},
\end{equation}
where ${d}_{n,m}$ represent the ray's direction of pixel $(n,m)$. The horizontal-plane distance $D_n$ to the origin then can be calculated as:
\vspace{-2pt}
\begin{equation}
D_n = \sqrt{x_n^{2} + z_n^{2} }.
\end{equation}
Concatenating the results from the four depth maps, the final $120 \times 1$ matrix ${D}=\{D_k \mid k=1,...,120\}$ representing the shortest distances across 120 intervals can be obtained.

Given the computed shortest distances \({D}\), we construct the occupancy mask \({M} \in \{0,1\}^{120 \times 12}\) by mapping these distances into a discrete 120 × 12 grid, ensuring alignment with the classifier's heatmap \(\mathbf{P}\), where angles are spaced every 3 degrees, and distances range from 0.25m to 3.00m in 0.25m increments.

For each angle index \( k \), we determine if the corresponding distance \( d_j \) is within the navigable range:

\begin{equation}
M_{k,j} =
\begin{cases}
1, & d_j \leq D_k \\
0, & d_j > D_k
\end{cases}
\end{equation}
where \( M_{k,j} = 1 \) indicates that the region at angle \( k \) and distance \( d_j \) is traversable, and \( M_{k,j} = 0 \) indicates that an obstacle blocks navigation beyond \( D_k \).

To enforce waypoint predictions that align with physically navigable regions, we introduce an occupancy-aware loss based on binary cross-entropy (BCE) loss between the predicted heatmap ${P} $ and the generated occupancy mask ${M}$:

\begin{equation}
L_{occ} = \text{BCE}(P, M)
\end{equation}

By minimizing this loss, the model discourages predictions in non-navigable areas while reinforcing confidence in traversable locations. This improves the accuracy and safety of waypoints generated in continuous environments.

Extend from original loss from Eq~\ref{eq:L_vis}, the final objective balances these two components: 
\begin{equation} L_{total} = L_{vis} + \lambda_{occ} *L_{occ} \end{equation} 
where $\lambda_{occ}$ control the relative importance of occupancy constraints. This combined loss ensures that the predicted waypoints are not only accurate but also feasible for real-world navigation.

\subsection{MLLM as Navigator}\label{sec:method_nav}

Given the following components: (1) the \textbf{Instruction} ($L$), (2) the \textbf{Observation} ($O_t$), which consists of the set of visual data for the navigable waypoints predicted by our waypoint predictor at each time step, (3) the \textbf{Action Space} ($A_t$), which includes discrete directional commands (e.g., ``turn left to\dots'') paired with their corresponding viewpoint observations or images, and (4) the \textbf{History} ($H_t$), a succinct record of the agent’s previous actions that maintains context across multiple navigation steps.
We carefully design a history-aware and spatio-temporal reasoning framework from two distinct perspectives to enhance the navigational proficiency of MLLMs: (1) a history-aware spatiotemporal single-expert prompt system and (2) an adaptive path planning procedure.

\subsubsection{History-aware Single-expert Prompt System }
Prior VLN approaches such as NavGPT~\cite{zhou2024navgpt} and DiscussNav~\cite{long2023discuss} rely on multiple specialized `experts' (e.g., for instruction parsing, historical summarization, or textual conversion of visual input), which not only risk cumulative information loss and redundant computation but are also limited to discrete environments. In contrast, our method builds upon MapGPT~\cite{chen2024mapgpt}’s single-expert paradigm and extends it to VLN-CE. Specifically, by unifying textual and visual processing within a single model, our agent can handle continuous spatial settings without requiring separate modules. As shown in Figure~\ref{fig:prompt}, at each time step ($t$) it processes all relevant prompts, including task description, instruction, history, observations, and action options, and outputs precisely one selected action, thereby consolidating the reasoning process and avoiding extraneous intermediate stages.

Given the inherent complexity of continuous environments, we systematically broaden the action space from two distinct perspectives to enhance adaptability in VLN-CE. First, we introduce multi-scale turning prompts. Instead of a single turn left action, we differentiate between ``turn slight left" ($\approx 30^\circ$) for slight directional shifts and 
``turn sharp left" ($\approx 90^\circ$) for large-angle rotations. This granularity improves the agent’s ability to align with navigable waypoints more precisely.
Secondly, we incorporate the RAM~\cite{Zhang2023RecognizeAA} model for fine-grained object recognition and classification in RGB images, thereby providing zero-shot semantic labeling for each detected object. This approach enriches the agent’s situational awareness of candidate waypoints. Let \({S} = \{s_{1}, s_{2}, \ldots, s_{n}\}\) denote the set of objects in the scene. The detection process analyzes the RGB images (candidate waypoints) to describe each object inside of the image, formally expressed as:
\begin{equation}
    s_{i} = \mathrm{RAM}(I_{\mathrm{rgb}}), \quad i=1,2,\ldots,n.
\end{equation}
Our full action space is shown in Figure~\ref{fig:prompt}-\textit{``Action Options"}.

History navigational awareness is crucial for avoiding repeated exploration and enabling strategic movement. As the limitations of history representations, as discussed in Sec.II, to address those limitations, we expanded the history information by extra scene description from previous action space that incorporated with thee RAM model. Our history is then expanded as shown in Figure~\ref{fig:prompt}-\textit{``History"}.

\subsubsection{Adaptive path planning with backtrack}

Prior LLM-based agent like Open-Nav\cite{opennav} navigator often operate with a local perspective by analysing the current candidate waypoints. We utilize the ability of MLLMs that can take advantage of multi-step planning, especially when the environment structure is available in image-textual form. This multi-step reasoning helps manage longer tasks involving complex or ambiguous instructions. To further enhance navigation performance, we introduce an adaptive path planning module wherein the model iteratively generates, updates, and refines a multi-step plan with backtrack mechanism. This plan encompasses potential exploration of candidate nodes, and importantly, incorporates backtracking strategies that allow the agent to revisit previous locations to correct errors and avoid redundant exploration. 

As shown in Figure~\ref{fig:prompt}-\textit{``Thought \& Action \& Backtrack"}, at each step, alongside the single chosen action, the model generates a short “thought” describing how it will proceed over subsequent moves. This plan may involve visiting several nodes, returning to previously explored areas, and reevaluating as new information is obtained. The plan  \(P_t\) becomes an input for the next step, meaning the model can revise its earlier intentions based on fresh observations or contextual details. If a chosen path does not yield the required landmarks or fails to match key instructions, the agent can revert to prior nodes by an extra action option at the end of action space: 

\textcolor{gray}{[`F. Move back to last position in an opposite direction']}

This mechanism is especially valuable in large or convoluted environments where mistakes or incomplete instructions often lead to dead ends.

\section{Experiments}

\subsection{Experiment Setup}
\noindent\textbf{Simulated Environments} Our implementation is built upon the Habitat simulator~\cite{savva2019habitat}. We evaluate our waypoint predictor on the MP3D~\cite{chang2017matterport3d} val-unseen set, strictly adhering to the evaluation protocol outlined in~\cite{hong2022bridging}. Furthermore, we assess the performance of our navigator by benchmarking it against several learning-based VLN methods as well as LLM/MLLM-based approaches, following the evaluation framework of Open-Nav~\cite{opennav} across 100 episodes to ensure direct comparability. The waypoint predictor is a learning-based module. For its training, we employ the AdamW optimizer~\cite{adamw} with a batch size of 128 and a learning rate of \(10^{-6}\), and we train the model for 300 epochs. The $\lambda_{occ}$ is set to 0.5. In contrast, the MLLM navigator is deployed via the gpt-4o-2024-08-06 API, which serves as our zero-shot inference backend.

\noindent\textbf{Real-world Environments}
We perform real-world experiments using a Turtlebot 4 mobile robot equipped with an OAK-D Pro camera mounted at a height of 70 cm. Our method relies exclusively on the camera for perception, without any contribution from external mapping modules. We evaluate our approach with and without the backtracking mechanism, and compare it against the pre-trained model proposed in~\cite{hong2022bridging}, deploying these systems on the Turtlebot 4 in a semantically rich indoor environment. The robot rotates in precise 30° increments, capturing 12 images to form a complete panoramic view that is subsequently processed by our navigation system. All methods are executed in real-time on a laptop equipped with an NVIDIA RTX 3080 Mobile GPU (16 GB VRAM). At each navigation step, the robot advances according to the predicted angle and direction. For evaluation, following the experimental protocol established in DiscussNav~\cite{long2023discuss}, we define 25 distinct navigation instructions that incorporate open-vocabulary landmarks (e.g., ``dual arm robot''), fine-grained landmark details, and scenarios involving multiple room transitions. This evaluation framework was designed to rigorously assess the generalizability and robustness of our approach relative to learning-based baselines.

\noindent\textbf{Evaluation Metrics}
We evaluate our waypoint predictors using four principal metrics. Specifically, $\lvert \Delta \rvert$ measures the discrepancy between the predicted number of waypoints and the ground-truth count, while \%Open represents the proportion of predicted waypoints that remain unobstructed by obstacles. $S_{way}$ represents the alignment between the model-predicted waypoints and the target heatmap. Further, $d_C$ denotes the Chamfer distance, and $d_H$ the Hausdorff distance, both widely employed for quantifying spatial discrepancies between point clouds. 

We assess our navigator against several learning-based VLN methods and LLM/MLLM-based approaches using standard VLN evaluation metrics: success rate (SR), oracle 
success rate (OSR), success weighted by path length (SPL), trajectory length (TL), navigation error (NE), and Collisions. Consistent with the evaluation protocol in Open-Nav~\cite{opennav}, an attempt is deemed successful if the agent halts within 3 meters of the target in the VLN-CE 
environment, whereas a stricter 2-meter threshold is applied for real-world scenarios. These criteria ensure a balanced and fair evaluation across diverse settings, accounting for both navigational precision and spatial constraints.

\subsection{Results in Simulating Environments}

Table~\ref{tab:table_performance_comparison} presents a comprehensive performance comparison of various models on the R2R-CE dataset in the Habitat simulator. The table is organized into two blocks: \emph{Supervised} approaches (upper block) and \emph{Zero-Shot} approaches (lower block). Supervised methods benefiting from domain-specific training, achieve higher success rates (52\% and 60\% respectively) and robust performance across additional metrics, thus holding an inherent advantage in simulated environments. In contrast, zero-shot models, which are deployed without any task-specific fine-tuning, generally exhibit lower SPL and SR. When evaluated strictly under zero-shot conditions (lower block), our zero-shot approach outperforms all other zero-shot methods, attaining an OSR of 51\%, an SR of 29\%, and an SPL of 22.46\%. These results highlight the superior waypoint alignment and decision-making of our method, which translates into enhanced goal-reaching capability and path efficiency. Overall, our findings confirm that leveraging generalizable representations in a zero-shot setting can significantly improve navigation performance, narrowing the gap with fully supervised models. Table \ref{tab:waypoint_pred} presents a comparison of different waypoint predictors on the MP3D Val-Unseen dataset. Our method achieves the highest \%Open (87.26\%), indicating better waypoint prediction in unobstructed areas. It also attains the lowest \(d_h\) (1.96), suggesting improved waypoint alignment.
\begin{table}[t]
\caption{Comparison on Simulated Environment in R2R-CE dataset}
\label{tab:table_performance_comparison}
\vspace{-15pt}
\begin{center}
\resizebox{\linewidth}{!}{
\begin{tabular}{l|ccccc}
\toprule
\textbf{Method} & \textbf{TL} & \textbf{NE}$\downarrow$  & \textbf{OSR}$\uparrow$ & \textbf{SR}$\uparrow$ & \textbf{SPL}$\uparrow$ \\
\midrule
\rowcolor{Grey!20}\multicolumn{6}{c}{\textbf{Supervised}}\\
CMA\cite{hong2022bridging}& 11.08 & 6.92  & 45 & 37 & 32.17 \\
RecBERT\cite{hong2022bridging}         & 11.06 & 5.8   & 57 & 48 & 43.22 \\
BEVBert\cite{an2023bevbert}               & 13.63 & 5.13 & 64 & 60 & 53.41 \\
ETPNav\cite{An2023ETPNavET}                 & 11.08 & 5.15 & 58 & 52 & 52.18 \\
\midrule
\rowcolor{Cerulean!20}\multicolumn{6}{c}{\textbf{Zero-Shot}} \\
Random                 & 8.15  & 8.63   & 12 & 2  & 1.50 \\
LXMERT\cite{tan2019lxmert} & 15.79 & 10.48  & 22 & 2  & 1.87 \\
MapGPT-CE-GPT4o~\cite{chen2024mapgpt} &12.63&8.16&21&7&5.04\\
DiscussNav-GPT4\cite{long2023discuss}             & 6.27 & 7.77   & 15 & 11 & 10.51 \\
Open-Nav-Llama3.1\cite{opennav}         & 8.07  & {7.25}  & 23 & 16 & 12.90 \\
Open-Nav-GPT4\cite{opennav}         & 7.68  & {6.70}  & {23} & {19} & {16.10} \\
\midrule
\textbf{Ours} & 13.09  & 7.01 & \textbf{51}& \textbf{29} &\textbf{22.46} \\
\bottomrule
\end{tabular}}
\vspace{-10pt}
\end{center}
\end{table}
\begin{table}[t]
    \caption{COMPARISON ON REAL-WORLD ENVIRONMENTS}
    \label{tab:real_robot_results}
    \vspace{-8pt}
    \centering
    \begin{tabular}{l|cc}
        \toprule
        \textbf{Method} & \textbf{SR$\uparrow$} & \textbf{NE$\downarrow$} \\
        \midrule
        RecBERT~\cite{hong2022bridging} & 20 & 4.36 \\
        \midrule
        Ours (w/o Backtrack) & 24 & 3.28 \\
        Ours (with Backtrack) & \textbf{36} & \textbf{3.06} \\
        \bottomrule
    \end{tabular}
        \vspace{-15pt}
\end{table}

\subsection{Results in Real-world Environments}
We also conduct experiments in real-world environments, as shown in Table~\ref{tab:real_robot_results}. For comparison, we select RecBERT~\cite{hong2022bridging}, which is the best-performing supervised training method for real-world tasks, as reported in~\cite{opennav}. 
Besides, 

BEVBert~\cite{an2023bevbert} and ETPNav~\cite{An2023ETPNavET} require the construction of pre-mapped real-world environments, making the comparison inequitable. 
Our method achieves higher performance in success rate than RecBERT. While pre-trained methods can offer advantages in simulated settings, these gains do not always translate to real-world tasks. By removing the need for extensive pre-training, our approach narrows the performance gap with fully supervised methods, and maintains superior interpretability and adaptability, rendering it more robust to unmodeled real-world perturbations. In addition, we compare our proposed method with and without the backtrack mechanism. Empirical results indicate that enabling backtrack yields consistently higher success rates and lower navigation errors. By allowing the agent to detect and recover from suboptimal waypoint selections, backtrack mitigates compounding mistakes. This ensures that even when the initial predicted route is imperfect, the system can efficiently refine its trajectory, leading to more reliable goal-reaching performance in complex, dynamic real-world scenarios. A demonstration of TurtleBot 4 navigation is included in the supplementary video.

\subsection{Ablation Study}
\noindent\textbf{Waypoint Predictor}
We conduct an ablation study to assess the impact of different components on waypoint prediction, as shown in Table~\ref{tab:ablation_waypoints}. The baseline~\cite{hong2022bridging} uses a ResNet-50 encoder with a batch size of 64, no cross-attention for RGBD fusion, and no occupancy-aware loss, achieving $S_{way}$ = 1.40 with 79.86\% of waypoints in navigable areas. Increasing the batch size to 128 slightly improves $S_{way}$ (1.41) but reduces \%Open, showing minimal benefits. Replacing ResNet-50 with Dinov2 enhances alignment ($S_{way}$ = 1.42) and \%Open (80.45\%), demonstrating the advantage of stronger feature representations. The best results occur when incorporating occupancy-aware loss, boosting $S_{way}$ to 1.48, \%Open to 87.26\%, and reducing spatial errors ($d_C$ = 1.03, $d_H$ = 1.96). These findings confirm that combining stronger encoders, attention mechanisms, and obstacle-aware losses enhances waypoint prediction quality.

\begin{table}[t]
    \centering
    \caption{Comparison of the waypoint predictors on MP3D Val-Unseen.\\
    $\left|\Delta\right|$: Waypoint discrepancy, \%Open: Predict in unobstructed area; $d_{C}$: Chamfer distance, $d_{h}$: Hausdorff distance}
    \vspace{-10pt}
    \label{tab:waypoint_pred}
    \resizebox{0.8\columnwidth}{!}{
    \begin{tabular}{lccccc}
        \toprule
       \multirow{2}{*}{\#} & \multirow{2}{*}{Model} & \multicolumn{4}{c}{MP3D Val-Unseen}\\
        \cmidrule(lr){3-6} & & $\left|\Delta\right|$ & \%Open $\uparrow$ & $d_{C}$ $\downarrow$& $d_{h}$ $\downarrow$ \\
        \midrule
        1 &Baseline & 1.37 & 80.18 & 1.08&2.16 \\
        2 &U-Net~\cite{u-net} & 1.21 & 52.54 & \textbf{1.01}&2.00 \\
        3 &RecBERT~\cite{hong2022bridging} & 1.40 & 79.86 & 1.07&2.00 \\
        4 &ETPNav~\cite{An2023ETPNavET} & 1.39  & 84.05 & 1.04 & 2.01\\
        
        \midrule
        5 &Ours & 1.41 & \textbf{87.26} & 1.03 & \textbf{1.96}\\
        \bottomrule
    \end{tabular}}
    \vspace{-5pt}
\end{table}

\begin{table}[t]
    \caption{Ablation study on different waypoint settings. 
    \\BS: Batch Size; C: CrossAttn; O: Occupancy-aware Loss
    }
    \vspace{-10pt}
    \label{tab:ablation_waypoints}
    \centering
    \resizebox{\columnwidth}{!}{
    \begin{tabular}{l c| c c | c c c c}
        \toprule
        \makecell[c]{RGB Encoder} & BS & C & O & $S_{\text{way}}$ ↑ & \%Open ↑ & $d_C$ ↓ & $d_H$ ↓ \\
        \midrule
        \makecell[c]{\multirow{2}{*}{ResNet-50}} & 64  & \ding{55}  & \ding{55}  & 1.40 & 79.86 & 1.07 & 2.00 \\
             & 128 & \ding{55}  & \ding{55}  & 1.41 & 79.28 & 1.08 & 2.04 \\
        \cmidrule(lr){1-8}
        \makecell[c]{\multirow{3}{*}{Dinov2}}  & 128 & \ding{55}  & \ding{55}  & 1.42 & 80.45 & 1.07 & 2.01 \\
          & 128 & \ding{51}  & \ding{55}  & 1.47 & 82.15 & 1.09 & 2.09 \\
          & 128 & \ding{51}  & \ding{51}  & \textbf{1.48} & \textbf{87.26} & \textbf{1.03} & \textbf{1.96}\\
        \bottomrule
    \end{tabular}}
    \vspace{-15pt}
\end{table}

\noindent\textbf{Navigator}
Table~\ref{tab:vln_ablation} presents an ablation study examining how our enhanced waypoint predictor and backtrack-enabled navigator complement each other in improving navigation. The baseline configuration (\#1). When only the enhanced waypoint predictor is introduced (\#3), NE is decreased to 7.11, SR is increased to 26\%, confirming that improved waypoint prediction can reduce misalignment and detours. By incorporating the Backtrack mechanism (\#2, \#4), we observe a substantial decrease in collision rate (from 0.067 to 0.044), underscoring the system’s enhanced ability to recover from errors during route exploration. When our full framework are activated (\#4), the system achieves its highest SR = 29\% and attains the best SPL = 22.46\%, indicating that efficient waypoint prediction and robust misstep recovery together yield superior navigational performance. These results highlight that while each proposed component individually boosts navigation accuracy and reliability, their combination provides a synergistic advantage, offering the most effective and collision-robust navigation strategy overall.

\begin{table}[t]
    \caption{Ablation study on enhanced waypoint predictor and backtrack policy.
    }
    \label{tab:vln_ablation}
    \vspace{-10pt}
    \centering
    \resizebox{\columnwidth}{!}{
    \begin{tabular}{c|c@{\hskip 2pt}c@{\hskip 2pt}|c@{\hskip 2pt}c@{\hskip 2pt}|ccccccc}
    \toprule
    \multirow{2}{*}{\#} & \multicolumn{2}{c}{\textbf{Waypoint Predictor}} & \multicolumn{2}{c}{\textbf{Navigator}}  & \multicolumn{6}{|c}{\textbf{R2R Dataset}}\\
    \cmidrule(r){2-5}
    \cmidrule(r){6-11}
    & Original & Enhanced & Original & Backtrack & TL& NE$\downarrow$ & OSR$\uparrow$& SR$\uparrow$& SPL$\uparrow$ & Collisions$\downarrow$\\
    \midrule
    1 & \cmark& & \cmark& &10.51 & 7.61& 36&24& 18.20& 0.067 \\
    2 & \cmark& & & \cmark& 12.75 & 7.11 & \textbf{53} & 28 & 21.22 & 0.044\\ 
    3 & & \cmark & \cmark& & 15.32 & 6.73 & 52 & 26 & 17.61 & 0.042\\ 
    4 & &\cmark & & \cmark& 13.09 & \textbf{7.01}& 51 & \textbf{29} & \textbf{22.46} & \textbf{0.033}\\ 
    \bottomrule
    \end{tabular}
    }
 \vspace{-15pt}
\end{table}


\section{Conclusion and future work}
In this paper, we propose a zero-shot VLN-CE framework that optimizes waypoint prediction and adaptive decision-making, achieves robust trajectory planning and efficient navigation errors recovering. Unlike previous approaches, our method enhances the waypoint predictor by re-designing its architecture: leveraging a robust vision encoder, a masked cross-attention fusion strategy, and an occupancy-aware loss function, yield more accurate and obstacle-avoiding predictions. We also conduct the first exploration of MLLMs as the navigator in VLN-CE and introduce a novel backtracking mechanism for MLLM-based VLN agents, demonstrating that our approach significantly narrows the performance gap with supervised methods and highlighting the impact of combining explicit spatial reasoning with the generalization prowess of MLLMs from extensive experiments in both simulated and real-world environments. Moving forward, we plan to extend our backtracking mechanism to support multi-step undo operations and enhance semantic understanding by incorporating visual-language segmentation and scene-aware prompts, which could further correct deeper navigation mistakes and improve navigation performance.

\addtolength{\textheight}{-12cm}   











\bibliographystyle{IEEEtran}
\bibliography{IEEEfull}

\end{document}